\pdfoutput=1

\documentclass[final,5p,times,twocolumn,authoryear]{elsarticle}

\usepackage{amssymb}
\usepackage{listings}
\usepackage{booktabs} 
\usepackage{adjustbox}
\usepackage{url}
\usepackage{hyperref}
\usepackage{multirow}
\usepackage{amsmath}
\usepackage{longtable}
\usepackage{enumitem}
\usepackage{graphicx}
\usepackage{subcaption}
\usepackage{xcolor}

\newcommand{\improve}[1]{\textsuperscript{\textcolor{green!60!black}{#1}}}

\journal{}

\begin{document}

\begin{frontmatter}



\title{Reinforcement Learning for Decision-Level Interception Prioritization in Drone Swarm Defense}


\author{Alessandro Palmas - Artificial Twin - \texttt{alex@artificialtwin.com}}

\begin{abstract}
The growing threat of low-cost kamikaze drone swarms poses a critical challenge to modern defense systems demanding rapid and strategic decision-making to prioritize interceptions across multiple effectors and high-value target zones. In this work, we present a case study demonstrating the practical advantages of reinforcement learning in addressing this challenge. We introduce a high-fidelity simulation environment that captures realistic operational constraints, within which a decision-level reinforcement learning agent learns to coordinate multiple effectors for optimal interception prioritization. Operating in a discrete action space, the agent selects which drone to engage per effector based on observed state features such as positions, classes, and effector status. We evaluate the learned policy against a handcrafted rule-based baseline across hundreds of simulated attack scenarios. The reinforcement learning based policy consistently achieves lower average damage and higher defensive efficiency in protecting critical zones. This case study highlights the potential of reinforcement learning as a strategic layer within defense architectures, enhancing resilience without displacing existing control systems. All code and simulation assets are publicly released for full reproducibility, and a video demonstration illustrates the policy's qualitative behavior.

\end{abstract}




\end{frontmatter}


\section{Introduction}
\label{introduction}

The widespread availability and affordability of commercial unmanned aerial vehicles (UAVs) has recently driven an unprecedented rise in their use across diverse domains, including logistics, inspection, surveillance, and agriculture. However, this rapid proliferation has also raised significant security concerns. In particular, coordinated drone swarms represent a growing threat to critical infrastructure, military installations, and high-value civilian targets. The ability of multiple autonomous drones to evade static defenses and overwhelm conventional response systems makes effective counter-swarm strategies a pressing research challenge.

\begin{figure}[h]
\vskip 0.2in
\begin{center}
\centerline{\includegraphics[width=\columnwidth]{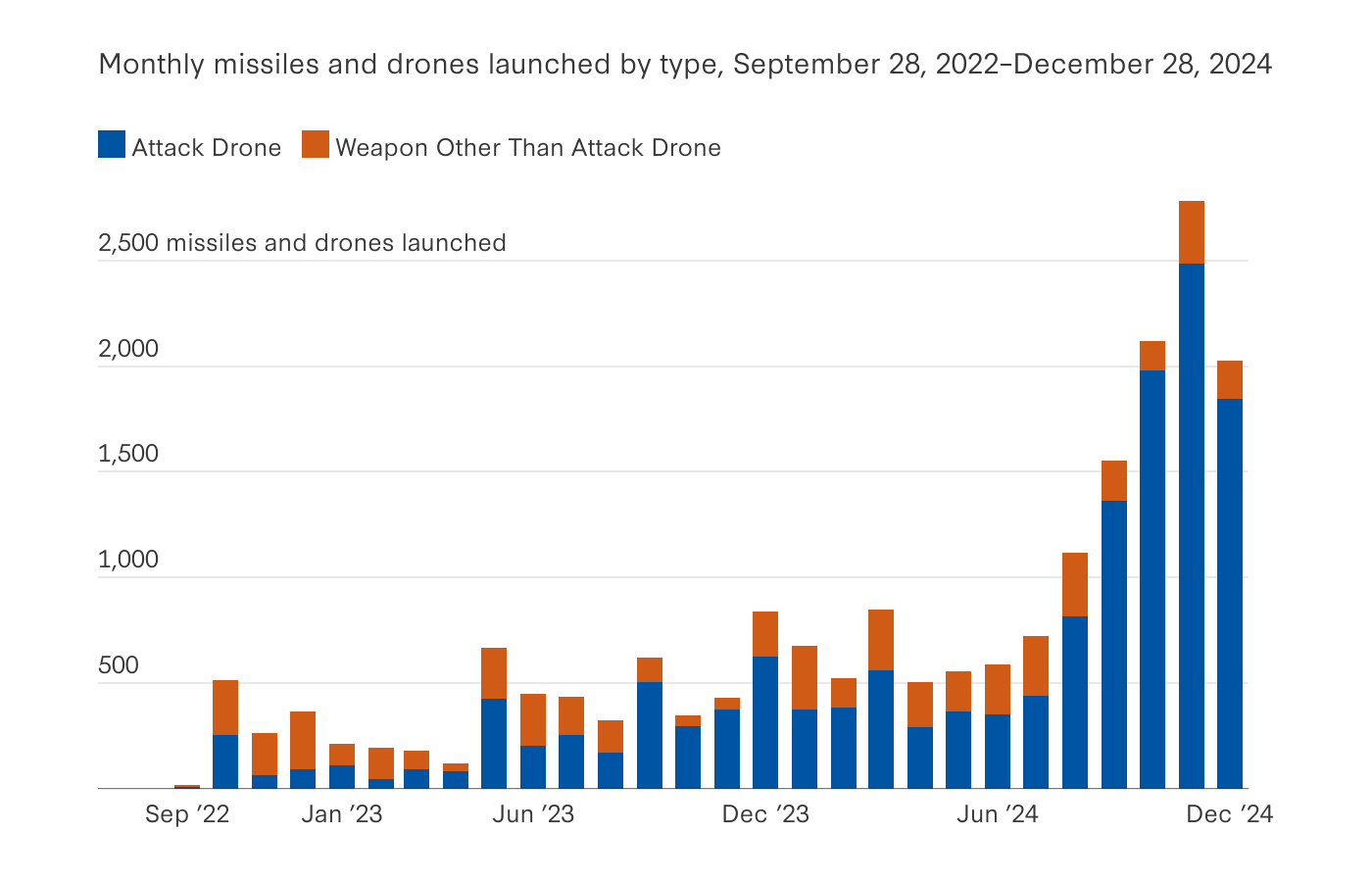}}
\caption{Increase in Drones Attack in Recent Conflicts - Source: \cite{csis2024}}
\label{drones_chart}
\end{center}
\vskip -0.2in
\end{figure}

A recent increase in reported drone-related incidents, including hostile flyovers, surveillance breaches, and attack attempts, highlights the urgency of this issue. Figure~\ref{drones_chart} illustrates the upward trend in documented drone attacks \citep{csis2024}, underscoring the need for robust defense mechanisms capable of operating under uncertainty, partial observability, and real-time constraints.

Recent advancements in onboard perception and control have significantly enhanced the autonomy and intelligence of small multirotor drones. For example, deep learning frameworks tailored for real-time visual processing on lightweight embedded systems have been proposed to enhance onboard situational awareness \citep{xiao2025, palmas2022uavcv}. In parallel, reinforcement learning (RL) and end-to-end policies have enabled coordinated behaviors such as flocking, pursuit, and area coverage in multirotor swarms \citep{arranz2023, batra2022}. As these capabilities mature, so too does the threat posed by adversarial UAV systems, particularly those operating in coordinated, self-organizing swarm formations. While beneficial in civilian and commercial contexts, these developments have also enabled increasingly sophisticated malicious use cases, ranging from autonomous surveillance and payload delivery to swarm-based saturation attacks on critical assets.

Traditional control and rule-based systems offer some degree of mitigation, but they often fail to generalize to the variability and unpredictability of real-world drone swarm scenarios. More importantly, they lack the capacity to adapt or learn from repeated exposure to such complex threat environments. In this context, reinforcement learning offers a promising alternative, learning optimal policies directly through interaction with high-fidelity simulators that replicate realistic engagement conditions.

Rather than controlling physical actuators or directing kinetic responses, this work investigates deep RL as a decision-support system. Specifically, we explore its role in threat prioritization, ranking individual drones based on their current and potential impact on protected assets. The resulting prioritization informs a downstream controller or human operator about which threats require immediate attention, augmenting perception and tracking data with an interpretable proxy for “dangerosity.”

We present a case study demonstrating the superior performance of deep RL in generating prioritization strategies compared to a classical heuristic-based policy. Trained entirely in simulation, the RL model learns to minimize the cumulative impact on sensitive zones in a realistic drone intrusion scenario. The results highlight the practicality, scalability, and safety of deploying RL-driven decision-support tools in cyber-physical defense systems.

The remainder of this paper is structured as follows. Section~\ref{relatedwork} reviews relevant background and related work on drone swarm defense and reinforcement learning in decision-support contexts. Section~\ref{problemformulation} describes the simulated environment, system architecture, and formalization of the control problem. Section~\ref{methodology} details the reinforcement learning formulation, including agent design, action space, and reward function. Section~\ref{evaluation} presents the experimental setup and performance comparisons between the learned policy and a classical rule-based baseline. Section~\ref{discussion} offers reflections on the role of RL in safety-critical systems. Finally, Section~\ref{conclusion} concludes the paper and outlines future work directions.

\section{Related Work}
\label{relatedwork}

The rapid proliferation of unmanned aerial vehicles has spurred a surge in research on autonomous defense systems capable of detecting, prioritizing, and neutralizing aerial threats, particularly in swarm-based attack scenarios. These efforts span multiple levels of abstraction, from low-level trajectory control to high-level decision-making, with reinforcement learning increasingly adopted as a powerful framework for adaptability and autonomy under uncertainty.

A substantial body of prior work has focused on control-level autonomy for UAV swarms. For example, \cite{batra2022} demonstrated decentralized multi-agent deep RL policies capable of flying quadrotors in formation and executing pursuit-evasion tasks, with zero-shot policy transfer from simulation to real hardware. Similarly, \cite{arranz2023} applied centralized RL based on proximal policy optimization (PPO) to assign surveillance and tracking duties among cooperative UAVs, demonstrating RL's potential for task distribution in persistent monitoring scenarios. These works highlight the feasibility of end-to-end learning for control, but focus primarily on executing low-level maneuvers and coordination within friendly UAV teams.

Other research has shifted toward adversarial contexts, where defensive agents must respond to malicious or non-cooperative UAVs. For instance, \cite{zhou2025} introduced a federated multi-agent RL framework to enable moving target defense (MTD) in UAV swarm networks under denial-of-service (DoS) attacks, using frequency hopping and leader-switching to thwart adversarial interference. \cite{xuan2022} investigated hierarchical multi-agent RL models simulating offensive and defensive UAV swarms engaged in coordinated confrontations, while \cite{zhao2021} explored multi-agent PPO (MAPPO) strategies for UAV dogfighting, emphasizing joint decision-making and resource allocation in contested airspace. These studies reinforce the growing recognition of RL’s ability to manage adversarial, multi-agent dynamics, though they often couple learning directly to physical control or assume full observability and homogeneous agent roles.

At a more strategic level, have applied RL to task assignment and mission-level planning. \cite{puentecastro2022} addressed path planning for UAV swarms tasked with area coverage using a centralized actor-critic model, while \cite{jung2024} combined edge AI with multi-agent learning to support adaptive decision-making in real-time swarm operations. While these contributions begin to address decision-making beyond trajectory-level commands, they often assume fully observable, static environments, failing to model the uncertainty and partial observability typical of real-world threat response scenarios.

The interpretability of RL-based decision-making has also received growing attention. For instance, \cite{cetin2024} applied Shapley additive explanations (SHAP) in counter-drone scenarios to analyze RL agent behavior, helping validate learned policies for real-time deployment. However, this line of work typically focuses on post-hoc transparency, rather than exploring architectural design choices that could inherently improve robustness or prioritization in dynamic scenarios.

In contrast to prior work, which largely focuses on direct UAV control or reactive defense in small-scale engagements, our study addresses a complementary and often underexplored challenge: centralized, high-level threat prioritization in swarm-based attacks. Rather than controlling effectors directly, our approach learns a policy that selects which hostile UAVs to prioritize at each timestep, based on evolving battlefield context, available defenses, and drone threat profiles. This formulation enables the RL agent to serve as a decision-support system, capable of operating under noisy, partial observations and adapting to heterogeneous attack patterns.

The simulated environment captures key aspects of real-world engagements, including probabilistic observations, zone-specific damage potential, heterogeneous drone types, and imperfect action execution. Within this setting, we demonstrate that our RL-based threat prioritization system consistently outperforms heuristic and rule-based baselines across several tactical metrics. This reinforces the utility of RL not merely as a control mechanism, but as a principled framework for learning adaptive defense strategies under uncertainty and dynamic adversarial conditions.

\section{Problem Formulation}
\label{problemformulation}

We consider a simulated defense scenario in which multiple kamikaze drones autonomously navigate toward high-value zones within a protected area, aiming to collide with and damage them. The area is defended by a set of kinetic effectors.

\subsection{Simulation Environment}

The environment is a three-dimensional domain $\mathcal{D}$ containing a swarm spawn volume $\mathcal{V}{\text{spawn}}$ for $N$ hostile drones, and a target volume $\mathcal{V}{\text{target}}$ that includes $Z$ sensitive static \textit{zones}. This space also includes $M$ defensive \textit{effectors} (e.g., kinetic interceptors or directed energy weapons) tasked with intercepting drones before impact. Effectors are modeled as state machines, each with separate kinematic and weapon states:

Kinematic:
\begin{itemize}[itemsep=1pt, before=\vspace{-6pt}]
\item Chasing: when retargeting their aim. Due to finite angular speed in azimuth and elevation, effectors must often pass through this state before locking onto a target.
\item Tracking: when locked on a target.
\end{itemize}

Weapon:
\begin{itemize}[itemsep=1pt, before=\vspace{-6pt}]
\item Ready: weapon is prepared to fire.
\item Firing: actively engaging a target.
\item Charging: in cooldown after firing, requiring time before becoming ready again.
\end{itemize}

And the following constraints apply for the transitions:
\begin{itemize}[itemsep=1pt, before=\vspace{-6pt}]
\item Firing is only possible when the effector is in the "tracking" kinematic state and the "ready" weapon state.
\item Recharge occurs independently of the kinematic state.
\end{itemize}

Each episode begins with a randomized swarm of hostile drones spawned within $\mathcal{V}_{\text{spawn}}$, each targeting zones based on predefined but unobservable rules. The defender's task is to \textit{prioritize which drones to intercept at each timestep}, considering constraints such as limited firing rates, angular velocity limits, and line-of-sight restrictions.

The simulation is discrete-time with fixed step size $dt$, multi-agent, and partially observable, providing noisy information from the defender’s perspective. It incorporates physics-informed drone trajectories (e.g., maximum speed) and realistic effector behavior (e.g., cooldowns, tracking limitations). At each timestep, the defender receives noisy state observations and must decide which drones to target. The environment supports large-scale swarm attacks and batch evaluation across hundreds of episodes.

\subsection{Threat Model}

Attackers are modeled as \textit{kamikaze drones}, autonomous agents programmed to reach and collide with one of the protected zones unless intercepted. Drones vary in speed, size, explosive power, and flight trajectory, including spawn point, intermediate waypoints, and target destination. Their behavior is pre-computed and non-adaptive, representing low-cost adversaries with increasing autonomous capabilities. A successful drone strike inflicts damage proportional to the target zone’s value and the drone’s explosive power. Due to limited effector availability, angular movement constraints, and firing delays, full protection is infeasible, making \textit{prioritization critical}.

\subsection{Prioritization Task}

The central decision problem is: at each timestep, given the current (partially observed) state, \textit{which drone should each effector target to minimize total damage during the episode}? This task becomes especially difficult with multiple simultaneous threats, each differing in urgency, distance, and potential damage. Poor prioritization can lead to catastrophic damage to critical assets. Operating under partial observability and time constraints, the defender must balance long-term outcomes against immediate risks.

\subsection{Environment Configuration}

Each element of the scenario is characterized by a set of configurable features for simulation customization.

\textbf{Sensitive zones}. Zones are assumed to be circular in shape and located on the ground at $z=0.0\ m$, and specified by the location of their center $\mathbf{c}_i$, radius $r_i$, and value $v_i$.

\textbf{Drones}. For each drone one has to specify maximum speed $w_j$, size $s_j$, possibly categorical (e.g. Small / Medium / Large), explosive power $p_j$, possibly categorical (e.g. Low / Medium / High), target coordinate $t_j$, trajectory $\tau_j$, assumed in the form of piecewise linear paths.

\textbf{Kinetic effectors}. Effectors are assumed having a static location but with a two degrees of freedom aiming system, azimuth and elevation. For each effector, one needs to specify location $\mathbf{e}_m$,  azimuth-elevation constraints $\mathcal{C}_m(\phi, \theta)$, maximum Az-El velocity $\dot{\phi}_{m\ \max}$, $\dot{\theta}_{m\ \max}$, and recharging time $T_{\text{recharge}}$. A neutralization probability model $P_{\text{hit}}(d)$ shared across all effectors is also required.

\textbf{Sensors}. The detection system is assumed to provide the noisy information on the state of the scenario. Its configuration is specified by: noise added to drone position detection $\epsilon_{\text{pos}}$ (e.g., Gaussian), size prediction accuracy $\mathbf{\epsilon}_{\text{size}}(s_j)$ (possibly probabilistic), explosive prediction accuracy $\mathbf{\epsilon}_{\text{power}}(p_j)$ (possibly probabilistic).

\begin{figure*}[t]
\vskip 0.2in
\begin{center}
\centerline{\includegraphics[width=\textwidth]{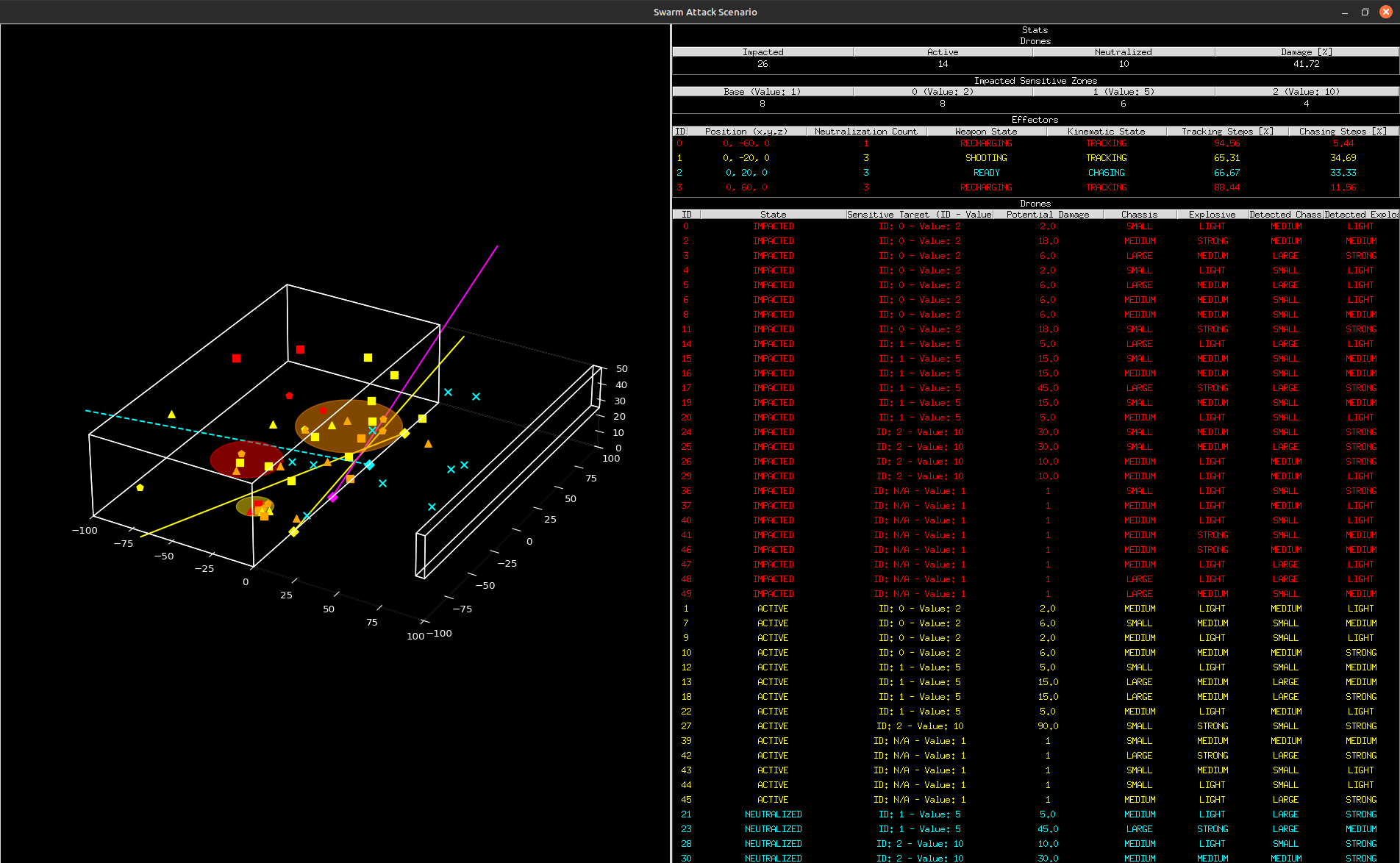}}
\caption{Simulation environment showing a snapshot of multiple kamikaze drones approaching protected zones and the corresponding effector system responses.}
\label{simulator}
\end{center}
\vskip -0.2in
\end{figure*}

A summary of all the configuration parameters can be found in Table~\ref{env_param} and a visualization of the environment is shown in Figure~\ref{simulator}.

\subsection{Classical Baseline Policies}

Several rule-based prioritization strategies are commonly used in real-world systems. Examples include:

\begin{itemize}[itemsep=1pt, before=\vspace{-6pt}]
    \item \textbf{Closest-first}: targets the drone with minimum distance to the effector.
    \item \textbf{Zone-weighted heuristic}: prioritizes drones heading toward zones with higher criticality, weighted by proximity.
    \item \textbf{Greedy minimization of expected loss}: Computes an urgency score for each threat as a function of distance-to-impact and zone value.
\end{itemize}

Although computationally efficient and interpretable, these policies \textit{fail to capture dynamic trade-offs and multi-agent interactions}, especially under high-density swarm attacks or adversarial strategies. They can be used as baseline comparators to evaluate the effectiveness of learned policies introduced in the following section.

\section{Methodology}
\label{methodology}

To address the challenge of adaptive and real-time effector coordination for swarm drone interception, we formulate the prioritization task as an episodic, discrete-time, partially observable Markov decision process (POMDP) and leverage deep reinforcement learning \citep{sutton1998} to learn an optimal policy through interaction with a high-fidelity simulation environment.

\subsection{Problem Formulation as Reinforcement Learning}

The RL agent operates at a fixed decision frequency and observes the current state of the simulation environment. The partial, noisy \textbf{observation space} encodes the following information:
\begin{itemize}[itemsep=1pt, before=\vspace{-6pt}]
    \item Drones position vector, type Box\footnote{\label{gymnasium}Following the standard Gymnasium API nomenclature \citep{towers2024}} with shape $3\times N$.
    \item Drones state, type MultiDiscrete\footnotemark[1] with shape $[3] \times N$
    \item Drones explosive, type MultiDiscrete\footnotemark[1] with shape $[3] \times N$
    \item Effectors Azimuth-Elevation angles, type Box\footnotemark[1] with shape $2\times M$ 
    \item Effectors kinematic state, type MultiBinary\footnotemark[1] with shape $M$ 
    \item Effectors weapon state, type MultiDiscrete\footnotemark[1] with shape $[3] \times M$
          
\end{itemize}

At each decision step, the agent selects one target drone per effector. The \textbf{action space} is therefore a MultiDiscrete\footnotemark[1] with shape $[N] \times M$.

The \textbf{reward function} returns a value of $0$ if no drone impacts a target zone. Otherwise, it returns the negative sum over all impacting drones of the product between each drone's explosive power and the value of the impacted zone.

An episode terminates when all drones have either impacted their targets or have been neutralized.

\subsection{Preprocessing and Normalization}

As standard practice in training neural networks, all observations are normalized. 

While not not applied to the MultiBinary observation, all the other categorical variables of type MultiDiscrete are normalized using standard one-hot encoding. 

Box-type observations are linearly scaled to fall within the interval $[-1.0, 0.0]$.

To provide temporal context, the drone position vectors from the previous $n$ steps are stacked along the observation dimension. This compensates for the absence of recurrence in the policy architecture and helps the agent infer motion trends.

Although drone impact points are initialized to be uniformly distributed across the target volume in each episode, variability in explosive payloads causes the total potential damage to vary between episodes. To address this, rewards are normalized by the theoretical maximum episode damage (i.e., if all drones reach their targets). This ensures that the total episode return remains within the range $[-1.0, 0.0]$.

\subsection{Learning Algorithm and Policy Architecture}

We use the proximal Policy Optimization Algorithm (PPO) \citep{schulman2017ppo} to train the agent, and compare its standard implementation with a variant incorporating action masking (MaskedPPO) \citep{huang2022}. The policy and value networks share a two-layer multilayer perceptron (MLP) backbone with 64 hidden units per layer, followed by ReLU activations. The policy head outputs a probability distribution over the discrete action space, while the value head estimates the expected return of the current state.

\subsection{Training Setup and Parameters}

Training is conducted using vectorized environments, with $32$ parallel simulation instances running in separate threads to accelerate data collection. The agent is trained for $80$ million steps using the Adam optimizer, with a learning rate linearly decayed from $2.5 \times 10^{-4}$ to $2.5 \times 10^{-6}$ and a discount factor of $\gamma = 0.998$.

The PPO clipping parameter is set to $0.15$ at the start of training and linearly decreased to $0.025$. Rollout length is set to $512$ steps, with $10$ training epochs per update and a batch size of $2048$.

A hyperparameter sweep was conducted to identify the most effective configuration. To ensure reproducibility, the complete training pipeline, environments, and configuration files are released as open source.

\section{Evaluation \& Results}
\label{evaluation}

The simulation environment was configured according to a specific setup, detailed in Table~\ref{spec_config}.

\begin{table}[h!]
\centering
\begin{tabular}{lp{6cm}}
\toprule
\multicolumn{2}{c}{Fixed: constant across episodes} \\
\midrule
$\mathcal{D}$ & $[-100,100]\times[-100,100]\times[0,50]$ m\\
$\mathcal{V}_{\text{target}}$ & $[-100,0]\times[-100,100]\times[0,50]$ m \\
$\mathcal{V}_{\text{spawn}}$ & $[95,100]\times[-100,100]\times[25,50]$ m \\
$dt$ & $0.1$ sec \\
$Z$ & 3 \\
$\{\mathbf{c}_i\}_{i=1}^{Z}$ & $[-30, -50, 0]$, $[-30, 50, 0]$, $[-60, -10, 0]$ m \\
$\{r_i\}_{i=1}^{Z}$ & 10, 30, 20 m \\
$\{v_i\}_{i=1}^{Z}$ & 2, 5, 10 \\
$N$ & 50 \\
$M$ & 4 \\
$\{\mathbf{e}_m\}_{m=1}^{M}$ & $[0, -60, 0]$, $[0, -20, 0]$, \\ 
& $[0, 20, 0]$, $[0, 60, 0]$ m \\
$\{\mathcal{C}_m(\phi, \theta)\}_{m=1}^{M}$ & $\phi \in [-\pi, \pi], \theta \in [0, \pi/2]$ \\
$\{\dot{\phi}_{m\ \max}\}_{m=1}^{M}$ & $\pi/2/s$ \\

$\{\dot{\theta}_{m\ \max}\}_{m=1}^{M}$ & $\pi/3/s$  \\
$\{T_{\text{recharge}}\}_{m=1}^{M}$ & 0.5 sec \\
$P_{\text{hit}}(d)$ & See figure~\ref{neutralization_prob} \\

$\epsilon_{\text{pos}}$ & $\mathcal{N}(0, \sigma^2)$, $\sigma_j=f(s_j)$, see table~\ref{pos_noise} \\
$\mathbf{\epsilon}_{\text{size}}(s_j)$ & See table~\ref{drones_size_classification} \\
$\mathbf{\epsilon}_{\text{power}}(p_j)$ & See table~\ref{drones_power_classification} \\
\midrule
\multicolumn{2}{c}{Variable: sampled at episode start} \\
\midrule
$\{w_j\}_{j=1}^{N}$ & See table~\ref{drones_features_distributions} \\
$\{s_j\}_{j=1}^{N}$ & See table~\ref{drones_features_distributions} \\
$\{p_j\}_{j=1}^{N}$ & See table~\ref{drones_features_distributions} \\
$\{t_j\}_{j=1}^{N}$ & Uniformly distributed across all zones, including the target volume \\
$\{\mathbf{\tau}_j\}_{j=1}^{N}$ & Piecewise linear with randomized intermediate waypoints \\
\bottomrule
\end{tabular}
\caption{Configuration parameters adopted for this study.}
\label{spec_config}
\end{table}

\begin{figure}[h]
\vskip 0.2in
\begin{center}
\centerline{\includegraphics[width=\columnwidth]{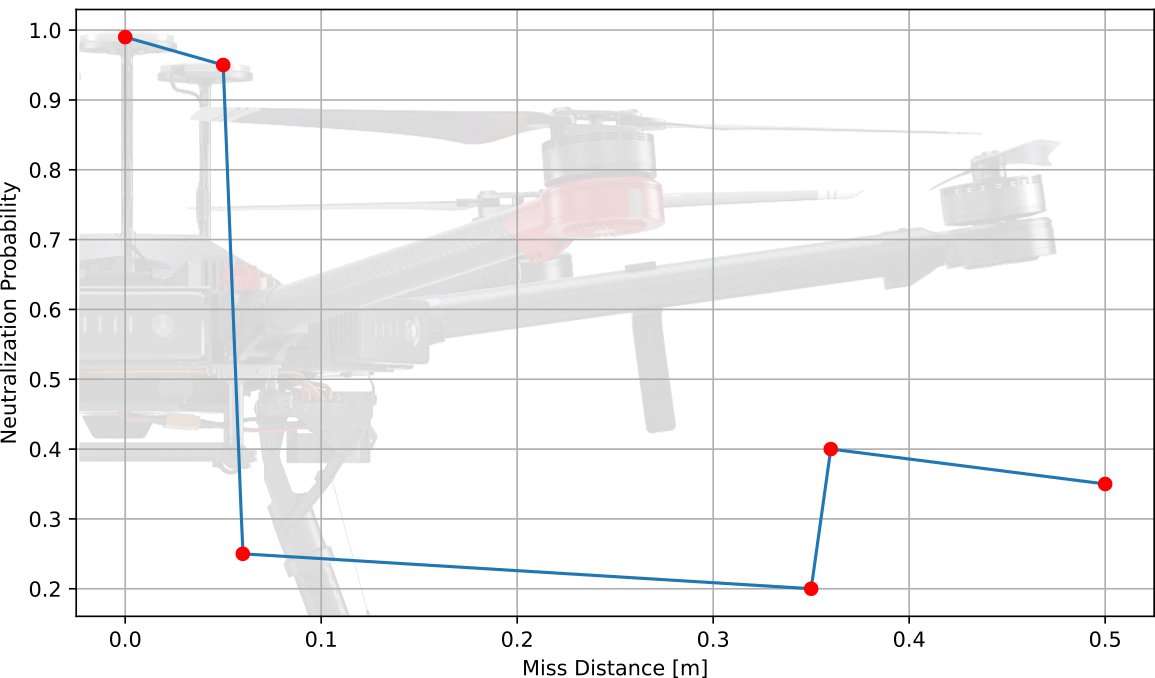}}
\caption{Neutralization Probability vs Miss Distance}
\label{neutralization_prob}
\end{center}
\vskip -0.2in
\end{figure}

The neutralization probability shown in Figure~\ref{neutralization_prob} is modeled as a piecewise linear function of the miss distance between an effector's aiming line and a drone’s actual position. Its shape reflects the typical architecture of multi-rotor drones, where both the central body and peripheral propellers are highly sensitive to impact.

The swarm is composed of drones varying in maximum speed, size, and explosive power across three distinct configurations.

\begin{table*}[h!]
\centering

\begin{subtable}[t]{0.48\textwidth}
\centering
\begin{tabular}{cccc}
\toprule
 & Small & Medium & Large \\
\midrule
$\sigma$ & 0.75 & 0.5 & 0.25 \\
\bottomrule
\end{tabular}
\caption{Position uncertainty as a function of size.}
\label{pos_noise}
\end{subtable}
\hfill
\begin{subtable}[t]{0.48\textwidth}
\centering
\begin{tabular}{ccccc}
\toprule
 & & & Detected & \\
 & & Small & Medium & Large \\
\midrule
 & Small & 0.8 & 0.1 & 0.1 \\
True & Medium & 0.1 & 0.8 & 0.1 \\
 & Large & 0.1 & 0.1 & 0.8 \\
\bottomrule
\end{tabular}
\caption{Drone size classification probability.}
\label{drones_size_classification}
\end{subtable}

\vspace{0.4cm}

\begin{subtable}[t]{0.48\textwidth}
\centering
\begin{tabular}{ccccc}
\toprule
 & & & Detected & \\
 & & Low & Medium & High \\
\midrule
 & Low & 0.8 & 0.1 & 0.1 \\
True & Medium & 0.3 & 0.4 & 0.3 \\
 & High & 0.1 & 0.2 & 0.7 \\
\bottomrule
\end{tabular}
\caption{Drone power classification probability.}
\label{drones_power_classification}
\end{subtable}
\hfill
\begin{subtable}[t]{0.48\textwidth}
\centering
\begin{tabular}{cccc}
\toprule
 & 10 m/s & 20 m/s & 30 m/s \\
\midrule
$w$ (max speed) & 0.4 & 0.4 & 0.2 \\
\midrule
\midrule
 & Small & Medium & Large \\
\midrule
$s$ (size) & 0.3 & 0.4 & 0.3 \\
\midrule
\midrule
 & Low & Medium & High \\
\midrule
$p$ (power) & 0.6 & 0.3 & 0.1 \\
\bottomrule
\end{tabular}
\caption{Probability distribution of drone features in the swarm.}
\label{drones_features_distributions}
\end{subtable}

\caption{Summary of uncertainty and classification probabilities in the simulated drone swarm environment.}
\label{combined_drones_tables}
\end{table*}

To assess the effectiveness of the learned RL policy, we conduct a thorough comparative evaluation against baseline strategies using a suite of randomized attack scenarios. The evaluation focuses on quantifying how well each policy minimizes damage to high-value zones under swarm attacks of varying intensity and configuration.

\subsection{Compared Policies}

We benchmark the following policies:
\begin{itemize}[itemsep=1pt, before=\vspace{-6pt}]
    \item \textbf{Random Policy}: Randomly selects a drone target for each effector at every decision step.
    \item \textbf{Classical Heuristic Policy}: A hand-crafted rule-based policy that prioritizes drones using a weighted combination of their explosive power, proximity to sensitive zones, and the importance of those zones. Defining $t$ the current timestep, $D = \{ d_1, \ldots, d_N \}$ the list of drones, $Z = \{ z_1, \ldots, z_M \}$ the list of sensitive zones, $\text{pos}(d_i^{t})$ the position of drone $d_i$ at time $t$, $\text{pos}(z_j)$ the position of zone $z_j$, $v_j$ and $r_j$ the value and radius of zone $z_j$, $e_i$ the explosive value of drone $d_i$ ($e_i \in \{1,2,3\}$ for low, medium, high power explosive respectively), $s_i$ the state value of drone $d_i$ at time $t$ ($s_i = 0$ for active drones, $s_i > 0$ for inactive drones), $d^{\text{max}}$ the maximum weighted distance (used for normalization purposes).

Then the distance score $S_i$ for drone $d_i$ is defined as:

\[
\begin{aligned}
w_i &= \sum_{j=1}^{M} \frac{ \left\| \text{pos}(z_j) - \text{pos}(d_i^{t}) \right\| }{v_j \cdot r_j} \\
w_i &= \frac{w_i}{e_i} + s_i \cdot 1000 \\
S_i &= \frac{ \min(w_i,\ d^{\text{max}}) }{0.5 \cdot d^{\text{max}}} - 1
\end{aligned}
\]    
    
    The policy code is included in the open sourced repository. 
    \item \textbf{RL Policy}: The policy learned via reinforcement learning, as described in Section~\ref{methodology}.
\end{itemize}

All policies are evaluated under identical simulation conditions and decision frequencies.

\subsection{Evaluation Setup}

Each policy is evaluated over $N = 100$ simulation episodes across five random seeds, with variations in:
\begin{itemize}[itemsep=1pt, before=\vspace{-6pt}]
    \item Initial drone positions.
    \item Drone characteristics: max speed, size and explosive power.
    \item Drones target points and flight paths.
\end{itemize}
as described in Section~\ref{methodology}.

Each scenario simulates a complete attack sequence, continuing until all drones are either neutralized or reach their targets. The evaluation compares the following performance metrics:

\begin{itemize}[itemsep=1pt, before=\vspace{-6pt}]
\item \textbf{Total Damage}: The weighted sum of impact events on zones, scaled by each zone’s criticality.
\item \textbf{Target Tracking Efficiency}: Assesses the policy’s ability to maintain high kinematic tracking performance.
\item \textbf{Weapon Utilization}: Measures how effectively the policy uses available interceptors.
\end{itemize}

\subsection{Results}

\begin{figure}[h]
\vskip 0.2in
\begin{center}
\centerline{\includegraphics[width=\columnwidth]{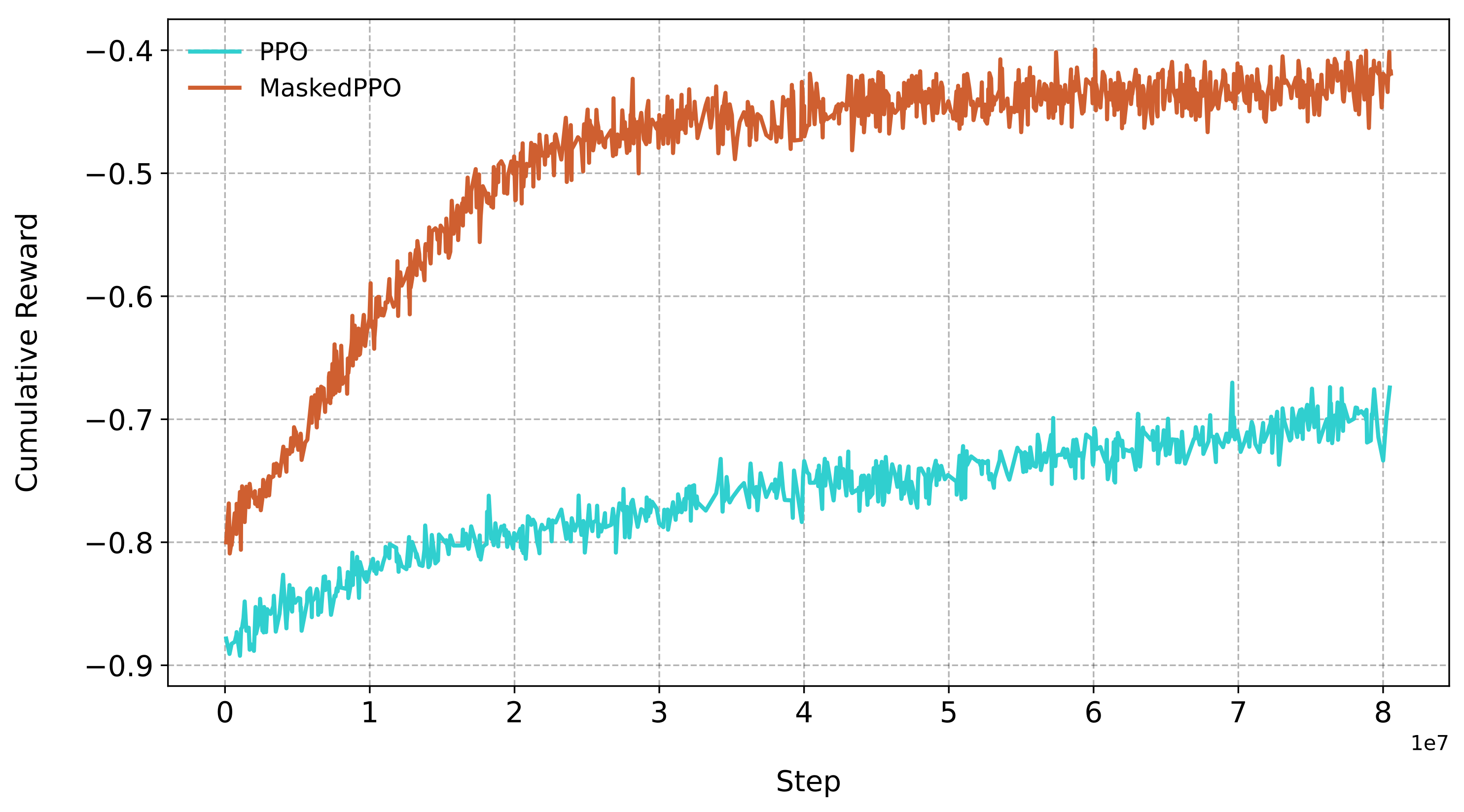}}
\caption{Training performance of PPO vs. MaskedPPO, showing cumulative reward per episode over environment steps. MaskedPPO converges ~10× faster by masking invalid actions (e.g., targeting already-neutralized drones), enabling more efficient and stable learning.}
\label{training_curves}
\end{center}
\vskip -0.2in
\end{figure}

All results in this section refer to the agent trained using the masked proximal policy optimization (MaskedPPO) algorithm. As illustrated in Figure~\ref{training_curves}, this variant significantly outperformed the standard PPO baseline during training, achieving convergence roughly 10 times faster in terms of sample efficiency. This improvement stems from the integration of action masking, a mechanism that dynamically excludes invalid actions, specifically removing, for each effector, any drone that has already been neutralized or has impacted a target. By pruning the action space, the agent avoids wasting capacity on irrelevant actions and focuses learning on valid threat prioritization. Given these benefits, the MaskedPPO-trained agent is adopted as the final DeepRL policy for all subsequent evaluations.

\begin{figure}[h]
\vskip 0.2in
\begin{center}
\centerline{\includegraphics[width=\columnwidth]{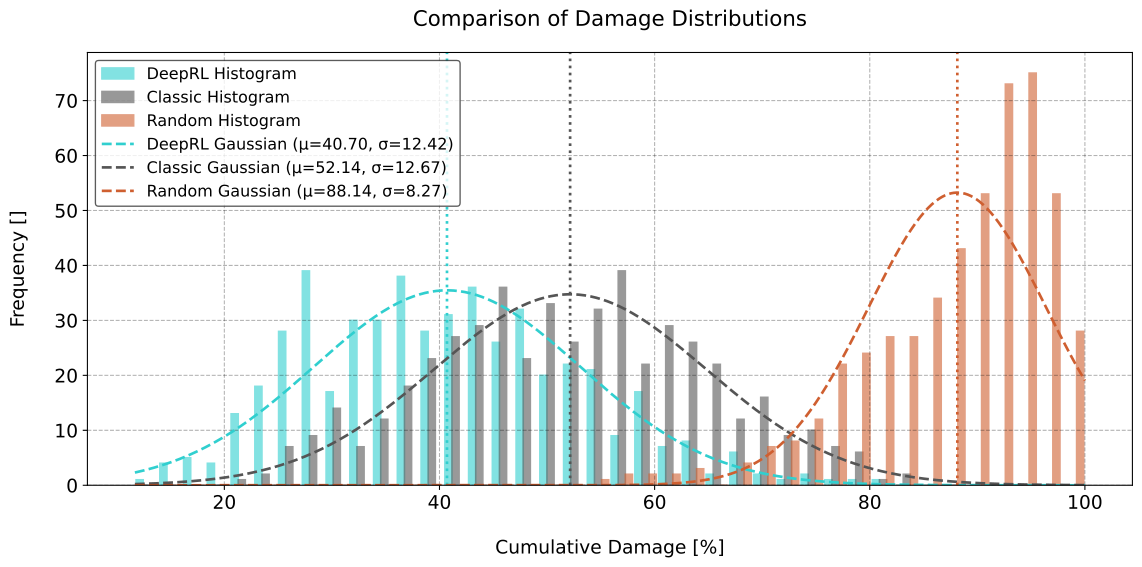}}
\caption{Distribution of total zone damage percentage for each controller. The RL agent consistently limits damage to critical zones compared to the heuristic baseline and random controller. }
\label{chart}
\end{center}
\vskip -0.2in
\end{figure}

\begin{figure*}[h]
\vskip 0.2in
\centering

\begin{subfigure}[b]{0.48\textwidth}
    \includegraphics[width=\linewidth]{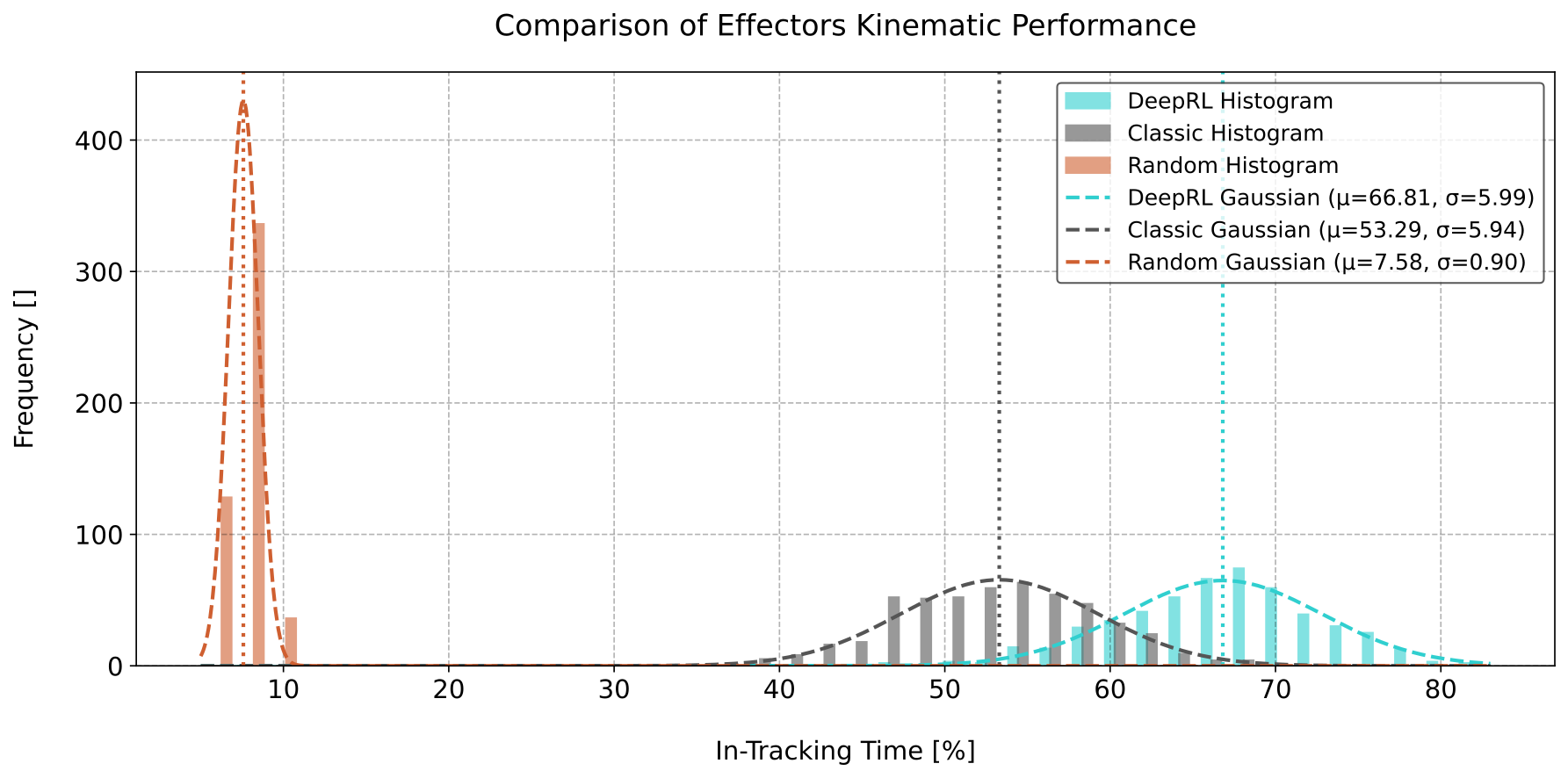}
    \caption{DeepRL vs Classic vs Random Controller - Target Tracking Efficiency Comparison}
    \label{fig:tracking_performance}
\end{subfigure}
\hfill
\begin{subfigure}[b]{0.48\textwidth}
    \includegraphics[width=\linewidth]{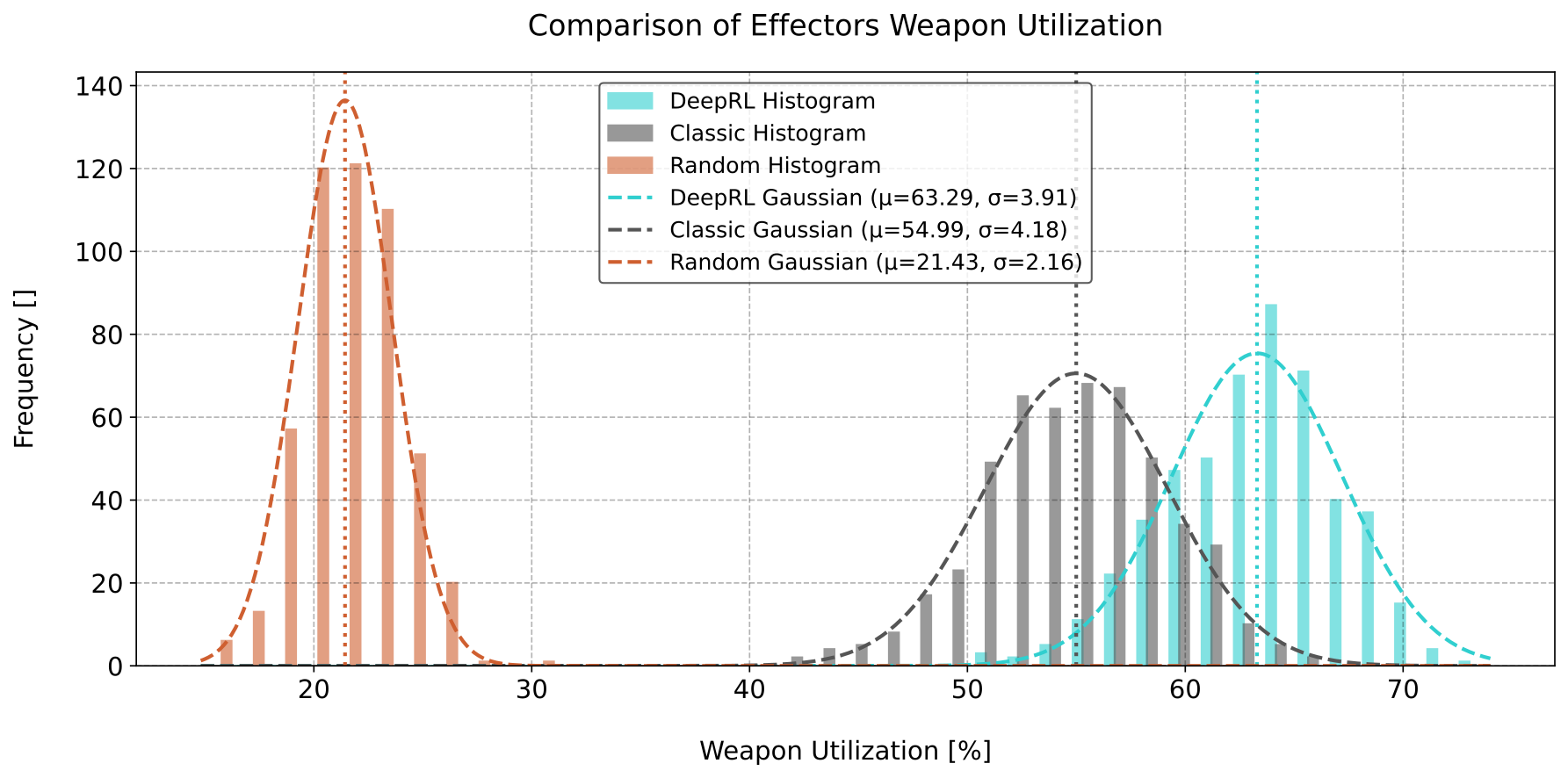}
    \caption{DeepRL vs Classic vs Random Controller - Weapon Utilization Comparison}
    \label{fig:weapon_utilization}
\end{subfigure}

\caption{Comparison of controller performance across two key enabling metrics: (a) target tracking efficiency and (b) weapon utilization. The DeepRL policy consistently achieves superior performance in both categories compared to the classical and random controllers, indicating improved resource allocation and sustained threat engagement over time.}
\label{fig:controller_comparison}
\vskip -0.2in
\end{figure*}

\begin{figure}[ht]
\vskip 0.2in
\begin{center}
\centerline{\includegraphics[width=\columnwidth]{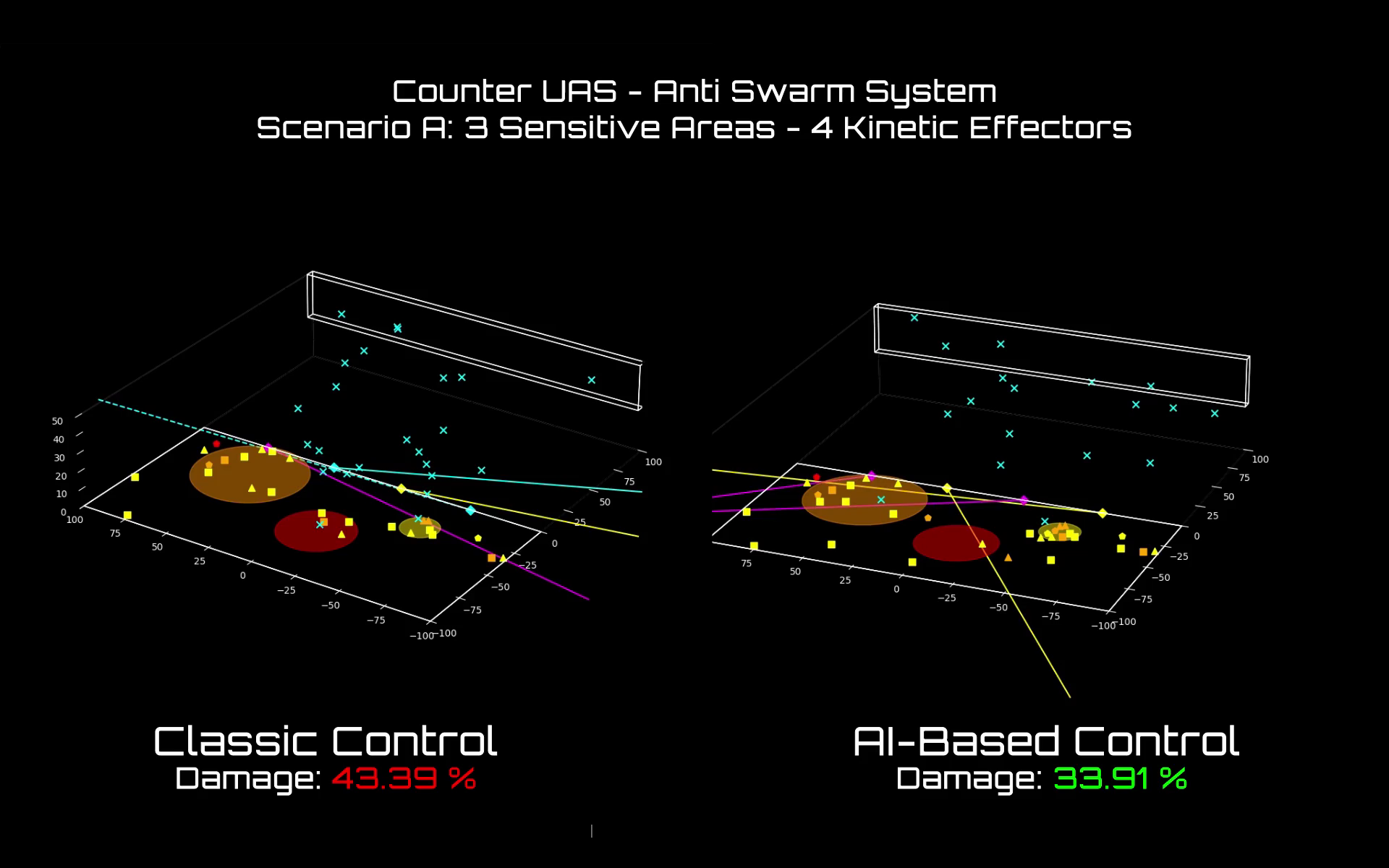}}
\caption{Snapshot of a final simulation step comparing the DeepRL and classic controllers. The DeepRL policy exhibits targeted threat neutralization around the high-value zone (red circle), prioritizing nearby or fast-approaching drones. In contrast, the heuristic controller allocates resources less efficiently, resulting in a more scattered defense and higher residual threat presence near critical assets.}
\label{final_state}
\end{center}
\vskip -0.2in
\end{figure}

As shown in Table~\ref{results_table} and Figure~\ref{chart}, Figure~\ref{fig:tracking_performance} and Figure~\ref{fig:weapon_utilization}, the RL policy significantly reduces average zone damage compared to both baselines, while also markedly improving tracking efficiency and weapon utilization. Figure~\ref{final_state} visually illustrates this effect: the high-value zone (red circle) is prioritized and protected, while lower-impact threats are deprioritized. The hand-crafted heuristic performs reasonably well but underperforms when compared to the RL policy due to its lack of adaptivity. Random policy performance is predictably poor, serving as a lower bound and sanity check.

Quantitatively, the RL policy achieves a $21.94$\% reduction in average damage compared to the heuristic baseline, alongside a $25.37$\% and $15.09$\% improvement in tracking efficiency and weapon utilization, respectively.

\begin{table}[h!]
\centering
\begin{tabular}{rcc}
\toprule
 & Classical & Reinforcement \\
 & Heuristic & Learning \\
  \midrule
 Total Damage (Avg) [\%] & 52.14 & \textbf{40.70}\improve{$\blacktriangledown22\%$} \\
In-Tracking Time (Avg) [\%] & 53.29 & \textbf{66.81}\improve{$\blacktriangle25\%$}  \\
Weapon Utilization (Avg) [\%]  & 54.99 & \textbf{63.29}\improve{$\blacktriangle15\%$}  \\
\bottomrule
\end{tabular}
\caption{Quantitative comparison between the classical heuristic and the RL policy over 500 simulated episodes (100 per seed × 5 seeds). The RL policy significantly outperforms the heuristic across all metrics, reducing average zone damage by nearly 22\%, and improving both tracking efficiency and weapon utilization.}
\label{results_table}
\end{table}

Intuitively, maximizing tracking time and weapon utilization contributes to damage reduction by increasing the number of interception opportunities.

\begin{figure*}[h]
\vskip 0.2in
\centering

\begin{subfigure}[b]{0.48\textwidth}
    \includegraphics[width=\linewidth]{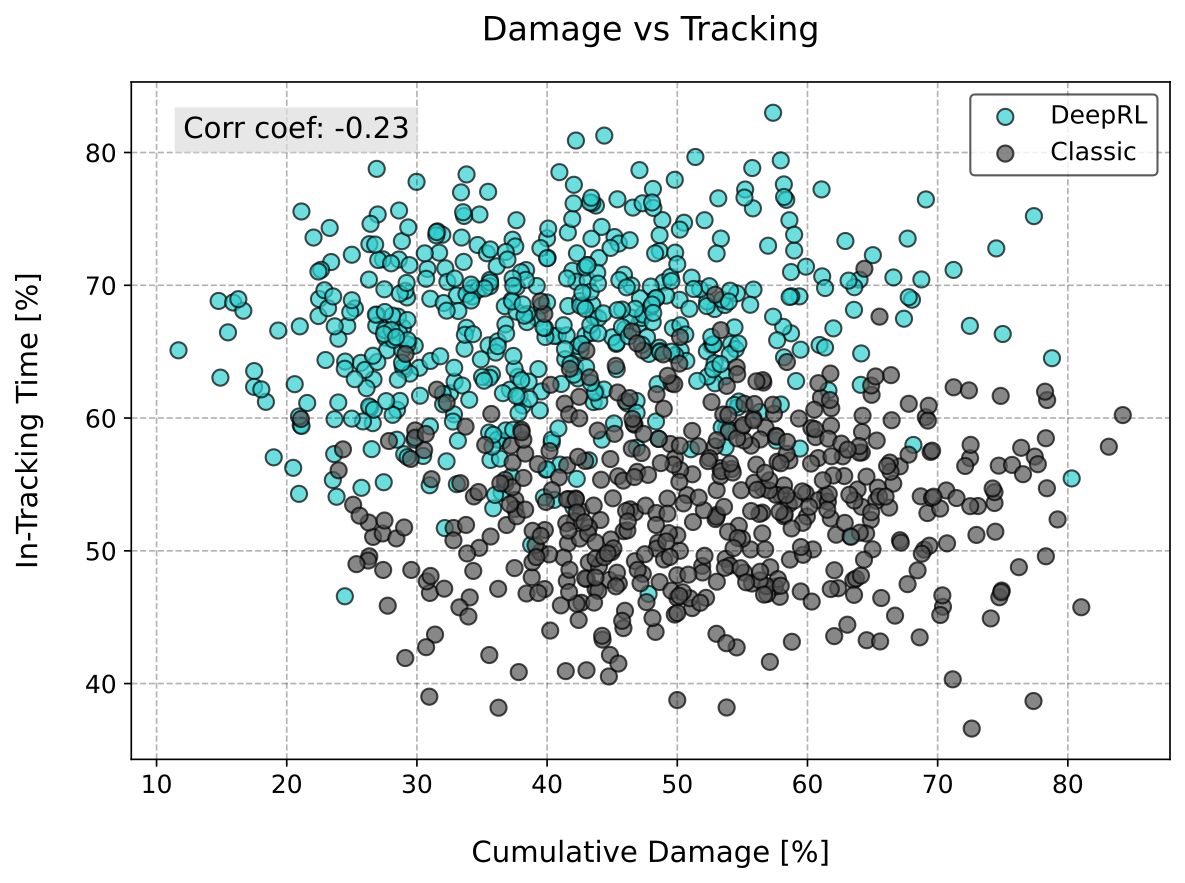}
    \caption{Damage - Tracking Correlation Plot}
    \label{fig:damage_tracking_correlation}
\end{subfigure}
\hfill
\begin{subfigure}[b]{0.48\textwidth}
    \includegraphics[width=\linewidth]{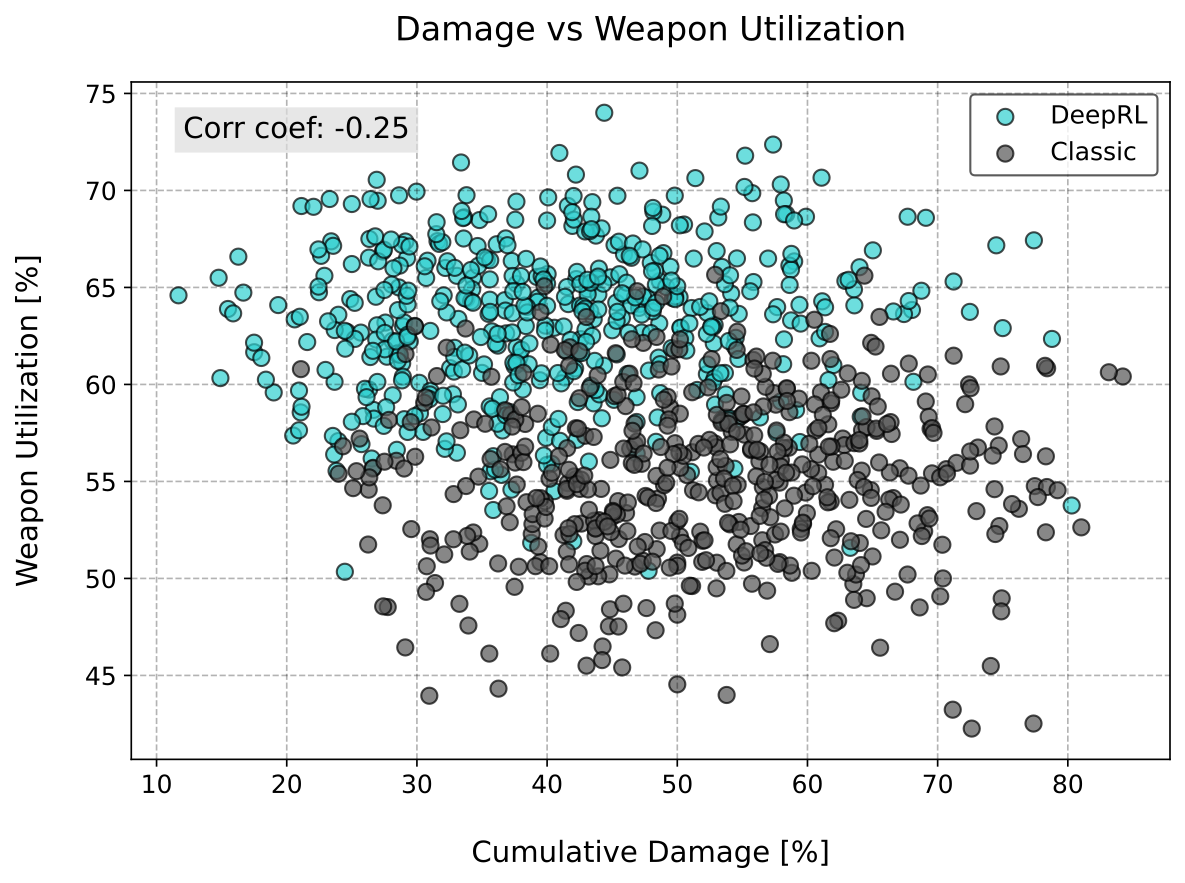}
    \caption{Damage - Weapon Utilization Correlation Plot}
    \label{fig:damage_weapon_usage_correlation}
\end{subfigure}

\caption{Scatter plots showing the relationship between zone damage and: (a) tracking efficiency, and (b) weapon utilization. While both correlations are negative, they are not strongly linear, highlighting that increased engagement opportunities (via better tracking and utilization) generally help reduce damage, but do not fully determine it due to the complex interplay of prioritization and threat behavior.}
\label{fig:correlations}
\vskip -0.2in
\end{figure*}

For completeness, Figures~\ref{fig:damage_tracking_correlation} and~\ref{fig:damage_weapon_usage_correlation} show the correlation between zone damage and the two metrics. As expected, the correlation is weak but negative in both cases.

\subsection{Interpretation}

The results confirm that the RL agent outperforms the classical baseline in protecting critical assets and exhibits more stable behavior under uncertainty. This robustness stems from the agent’s ability to learn adaptive prioritization strategies that consider long-term threat impact, rather than relying solely on proximity or fixed rules.

A supplementary video\footnote{Available at: \url{https://youtu.be/GooNFDk42Nw}} presents side-by-side simulations, highlighting emergent behaviors of the RL agent such as preemptive threat interception, effector load balancing, and zone-focused defense.

\section{Discussion \& Reflections}
\label{discussion}

The results highlight the potential of reinforcement learning as a powerful approach to address complex coordination problems in time-critical, high-stakes scenarios such as drone swarm defense. Several key insights emerge from our study.

\subsection{Strengths of RL in Multi-Threat Coordination}

A central strength of the RL policy lies in its ability to reason over high-level dynamics and prioritize threats in a manner that balances both urgency and strategic importance. Unlike rule-based policies that often collapse under the combinatorial complexity of real-world scenarios, the learned agent demonstrates:

\begin{itemize}[itemsep=1pt, before=\vspace{-6pt}]
    \item \textbf{Coordinated action selection}: efficiently allocating effectors across space and time.
    \item \textbf{Adaptability}: responding quickly to variations in attacker trajectories, densities, and layouts without retraining.
    \item \textbf{Emergent strategic behavior}: such as focusing on bottlenecks or sacrificial zones in order to shield high-value targets.
\end{itemize}

These capabilities arise not from manually encoding domain knowledge, but from exposing the agent to a sufficiently rich distribution of simulated encounters during training.

\subsection{Abstraction from Low-Level Dynamics}

A deliberate design decision was to frame the problem at the decision-making layer, abstracting away from the low-level control or kinematics of drones and effectors. This abstraction simplifies both training and deployment, enabling the learned agent to generalize across hardware and platform types, and to focus on threat prioritization rather than fine-grained physical execution.

In practice, this decouples high-level cognitive functions (e.g., which drone to neutralize) from low-level actuation (e.g., how to intercept), aligning well with how most Command and Control (C2) systems are architected.

\subsection{Human-in-the-Loop and Decision Support}

It is important to emphasize that the RL agent is not intended to operate as a fully autonomous controller. Instead, it serves as a \textit{decision-support tool}, providing real-time threat prioritization scores or ranked target lists for human operators.

Such integration supports scalable and explainable workflows in which the human remains in the loop, either approving recommendations or modifying them in real-time. This framework enhances trust, accountability, and regulatory compliance, critical factors in defense applications.

\subsection{Augmenting Existing C2 Pipelines}

The proposed system can be seamlessly integrated into existing C2 pipelines as an advisory module. At each decision step, the RL agent receives the current system state and outputs prioritized target allocations, which can be used to:

\begin{itemize}[itemsep=1pt, before=\vspace{-6pt}]
    \item Improve effector dispatching strategies
    \item Assist operators during high workload conditions
    \item Perform rapid “what-if” simulations for dynamic mission planning
\end{itemize}

This modularity enables incremental deployment and evaluation without replacing existing systems.

\section{Conclusion \& Future Work}
\label{conclusion}

This work introduced a reinforcement learning approach to prioritizing hostile drones in a simulated defense scenario. By framing the task as a high-level decision-making problem, abstracted from low-level control dynamics, we developed a policy that learns to allocate limited defensive resources in real time, minimizing damage to critical zones.

Our findings show that the RL-based agent consistently outperforms classical heuristic baselines in terms of average damage and policy stability. Notably, the agent demonstrates emergent coordination strategies and adaptability to varying threat patterns, without being explicitly programmed to do so.

\subsection{Future Directions}
While the current agent performs well in simulation, further work is required to enable real-world deployment:
\begin{itemize}[itemsep=1pt, before=\vspace{-6pt}]
    \item Develop domain adaptation techniques to enhance robustness against sensor noise and dynamic uncertainties.
    \item Design additional interpretability mechanisms to clarify the agent’s decision-making process.
    \item Extend to multi-agent settings where multiple defenders coordinate under shared or partial observations.
    \item Incorporate diverse threat types, such as decoys, stealth drones, or coordinated group behaviors, by expanding the simulator to better stress-test prioritization strategies.
    \item Evaluate robustness under partial observability by introducing sensor noise and detection gaps, and explore architectures such as recurrent policies or belief-state models.
    \item Progress toward field validation by integrating the RL agent into a closed-loop testbed or live trial environment, enabling real-time collaboration with human operators and legacy C2 systems.
\end{itemize}

Collectively, these steps aim to further validate and mature the proposed approach, supporting the development of trustworthy AI-driven decision-support systems for real-world defense applications.

\section{Resources}

To foster reproducibility and encourage further research, we provide access to the key assets used in this study:

\begin{itemize}
    \item \textbf{Code repository:} implementation of the simulation environment, training pipeline, evaluation tools, trained agent checkpoints, and classic baseline policy.\\
    \url{https://github.com/alexpalms/deeprl-counter-uav-swarm}
    
    \item \textbf{Demonstration video:} sample scenarios and visual comparisons between baseline and RL policies.\\
    \url{https://youtu.be/GooNFDk42Nw}
\end{itemize}

All resources will be maintained for at least 3 years after publication. For inquiries or collaboration, please contact the corresponding author.

\clearpage




\appendix

\onecolumn
\section{Simulation Environment Parameters}
\label{simulationparam}
\begin{longtable}{@{}lllp{7cm}@{}}
\caption{Simulation Environment Parameters} \\
\toprule
\textbf{Group} & \textbf{Variable Name} & \textbf{Symbol} & \textbf{Description} \\
\midrule
\endfirsthead
\toprule
\textbf{Group} & \textbf{Variable Name} & \textbf{Symbol} & \textbf{Description} \\
\midrule
\endhead
\bottomrule
\endfoot

\multicolumn{4}{@{}l}{\textit{1. Scenario}} \\
& Domain Bounding Box & $\mathcal{D}$ & Size of the simulated 3D domain \\
& Target Zone Volume Bounding Box & $\mathcal{V}_{\text{target}}$ & Region where the high-value target zones to be defended are located \\
& Drones Spawn Volume Bounding Box & $\mathcal{V}_{\text{spawn}}$ & Region where drones spawn \\
& Integration Time & $dt$ & Constant timestep for the simulation \\
\\
\multicolumn{4}{@{}l}{\textit{3. Sensitive Zones}} \\
& Number of Sensitive Zones & $Z$ & Total number of high-values target zones\\
& Sensitive Zone Centers & $\{\mathbf{c}_i\}_{i=1}^{Z}$ & Centers of each sensitive zone within the target volume \\
& Sensitive Zone Radii & $\{r_i\}_{i=1}^{Z}$ & Radius for each sensitive zone \\
& Sensitive Zone Values & $\{v_i\}_{i=1}^{Z}$ & Value of each sensitive zone \\
\\
\multicolumn{4}{@{}l}{\textit{2. Drone Swarm}} \\
& Number of Drones & $N$ & Total number of incoming drones in the swarm \\
& Drone Speed & $\{w_j\}_{j=1}^{N}$ & Flight speed per drone \\
& Size Category & $\{s_j\}_{j=1}^{N}$ & Physical size per drone; possibly categorical (e.g., S/M/L) \\
& Explosive Power Category & $\{p_j\}_{j=1}^{N}$ & Payload damage potential per drone; may be categorical (e.g., L/M/H) \\
& Drone Target Point & $\{t_j\}_{j=1}^{N}$ & Target point per drone \\
& Drone Trajectories & $\{\mathbf{\tau}_j\}_{j=1}^{N}$ & Piecewise linear paths from spawn point to target point, per drone \\
\\
\multicolumn{4}{@{}l}{\textit{3. Effectors}} \\
& Number of Kinetic Effectors & $M$ & Number of active effectors in the environment \\
& Effector Positions & $\{\mathbf{e}_m\}_{m=1}^{M}$ & Location of each effector in the domain \\
& Azimuthal-Elevation Constraints & $\{\mathcal{C}_m(\phi, \theta)\}_{m=1}^{M}$ & Bounds for horizontal and vertical rotation \\
& Max Azimuthal Velocity & $\{\dot{\phi}_{m\ \max}\}_{m=1}^{M}$ & Maximum horizontal rotation speed (rad/s) \\
& Max Elevation Velocity & $\{\dot{\theta}_{m\ \max}\}_{m=1}^{M}$ & Maximum vertical rotation speed (rad/s) \\
& Recharging Time & $\{T_{\text{recharge}}\}_{m=1}^{M}$ & Time delay between successive actions \\
& Neutralization Probability Model & $P_{\text{hit}}(d)$ & Probability as a function of miss distance $d$ \\
\\
\multicolumn{4}{@{}l}{\textit{4. Sensors}} \\
& Position Detection Noise & $\epsilon_{\text{pos}}$ & Positional noise added to drone detection (e.g., Gaussian) \\
& Size Category Classification Accuracy & $\mathbf{\epsilon}_{\text{size}}(s_j)$ & Model describing sensors size prediction accuracy; possibly probabilistic \\
& Explosive Category Classification Accuracy & $\mathbf{\epsilon}_{\text{power}}(p_j)$ & Model describing sensors explosive prediction accuracy; possibly probabilistic \\
\label{env_param}
\end{longtable}
\twocolumn


\newpage
  \bibliographystyle{elsarticle-harv} 
  \bibliography{rlCuas}


%
%
%
\end{document}